\documentclass[letterpaper,journal]{IEEEtran}
\usepackage{amsmath,amsfonts}
\usepackage{algorithmic}
\usepackage{algorithm}
\usepackage{array}
\usepackage[caption=false,font=normalsize,labelfont=sf,textfont=sf]{subfig}
\usepackage{textcomp}
\usepackage{stfloats}
\usepackage{url}
\usepackage{verbatim}
\usepackage{graphicx}
\usepackage{cite}
\usepackage{times}
\usepackage{multirow}
\usepackage{multicol}
\usepackage{makecell}
\usepackage{booktabs}
\usepackage{balance}
\usepackage[table]{xcolor}
\usepackage[colorlinks, linkcolor=white, implicit=false]{hyperref}
\begin{document}

\title{OCCA-Seg: Object and Channel Context Aggregation for Semantic Segmentation of Remote Sensing Images}

\author{Gang Yang\href{https://orcid.org/0000-0001-6096-9401}{\includegraphics[scale=0.08]{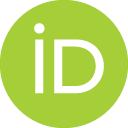}},
        Si Chen,
        Quan Zhou,
        Jiangchuan Li
\thanks{The authors are with Beijing Key Laboratory of Urban Spatial Intelligent Sensing and Digital Governance,
 Beijing Institute of Survey and Mapping, Beijing 100038, China (E-mail: yanggang3510@gmail.com) }
\thanks{Manuscript received April 19, 2021; revised August 16, 2021.}}

\markboth{Journal of \LaTeX\ Class Files,~Vol.~14, No.~8, August~2021}%
{Shell \MakeLowercase{\textit{et al.}}: A Sample Article Using IEEEtran.cls for IEEE Journals}


\maketitle

\begin{abstract}
Semantic segmentation, which assigns a class label to each pixel, is a key task in computer vision. 
However, semantic segmentation of remote sensing images (RSIs) remains challenging because of complex scenes and objects at multiple scales, which limits segmentation accuracy. 
To address these challenges, we propose a context aggregation scheme called object and channel context aggregation (OCCA), consisting of three components: 
object contextual attention module (OCAM), channel contextual attention module (CCAM), and aggregation module (AM). 
Firstly, motivated by the observation that each pixel shares the class label of the object to which it belongs, object-level context represents each pixel through aggregating context from its corresponding object class. 
Secondly, because feature channels are correlated with object categories, channel-wise context models channel-wise contextual dependencies conditioned on object categories. 
Finally, the context information from OCAM and CCAM are fed into AM and used to weight and modulate the object context, enhancing the discriminative ability of the segmentation model.
The experiments on several public remote sensing datasets including LoveDA, UAVid, ISPRS Potsdam, and Vaihingen datasets demonstrate
competitive improvements over existing methods. 
The implementation is available at \textcolor{blue!70!black}{\url{https://github.com/kuazhangxiaoai/OCCA-Seg}}.
\end{abstract}


\section{Introduction}

\IEEEPARstart{T}{he} semantic segmentation of remote sensing images (SSRSIs) is a fundamental task in
computer vision and plays an important role in a range of scene-understanding applications\cite{Ke2026ImprovedUNet}.
It supports geospatial applications such as urban planning, precision agriculture, and disaster
management. By assigning a class label to each pixel in aerial or satellite imagery, semantic
segmentation extracts structured information that can inform decision-making in urban design,
ecosystem monitoring, infrastructure development, etc. As geospatial information technologies
advance, the demand for SSRSIs continues to grow.
\par In recent years, deep convolutional neural networks (CNNs) have made remarkable progress in SSRSIs. 
Since the fully convolutional network (FCN) \cite{Long_2015_CVPR} was first proposed, extensive research has sought to improve
the performance of remote-sensing image segmentation. However, most CNN-based models suffer from the limited receptive fields. 
This limitation makes it difficult for the CNN-based models to obtain the long-range contextual information 
and understand the complex scene in the remote sensing images.
\par The limitation of receptive field originates from the inherent architecture of CNNs, 
which focus on local neighborhoods rather than long-range context information.
Due to the lack of long-range contextual information, these CNN-based methods are prone to 
misclassify pixels that belong to the same object but are spatially separated, especially 
in complex scenes with various object categories. Thus, the segmentation may be fragmented 
and the accuracy of object boundary identification may be reduced. 
\par In order to address the limited receptive field of CNNs and capture the long-range contextual information, 
we introduce the attention aggregation mechanism into the neural networks design.
In our proposed method, we design a novel context aggregation scheme called object and channel context aggregation (OCCA), 
which consists of three components: object contextual attention module (OCAM), channel contextual attention module (CCAM), and the aggregation module (AM). 
The OCAM captures the long-range contextual information by the features of all 
pixels that belong to the same object category.  
The CCAM models the channel-wise contextual dependencies conditioned on object categories, 
which enhances the discriminative representation of the object context.
The AM is responsible for integrating the attention information from both the OCAM and CCAM, 
thereby enhancing the overall feature representation and improving the segmentation performance.
\par In summary, this research proposes OCCA-Seg, combining object-level context from OCAM with channel-wise context from the CCAM.
It effectively integrates both types of context by AM to produce more discriminative feature representations.
This helps overcome the limitation of CNNs in capturing long-range contextual information.
We evaluate the proposed OCCA-Seg through extensive experiments on four benchmark datasets for semantic segmentation of remote sensing images, including ISPRS Potsdam, ISPRS Vaihingen, UAVid, and LoveDA dataset.
The mIoU of the OCCA-Seg is 87.51\%, 84.03\%, 68.9\%, and 55.5\% on the ISPRS Potsdam, ISPRS Vaihingen, UAVid, and LoveDA datasets, respectively.
These experiments verified the competitiveness of the OCCA-Seg among the state-of-the-art methods for semantic segmentation of remote sensing images.
\section{Related Work}
\begin{figure*} 
    \centering
    \includegraphics[width=18cm]{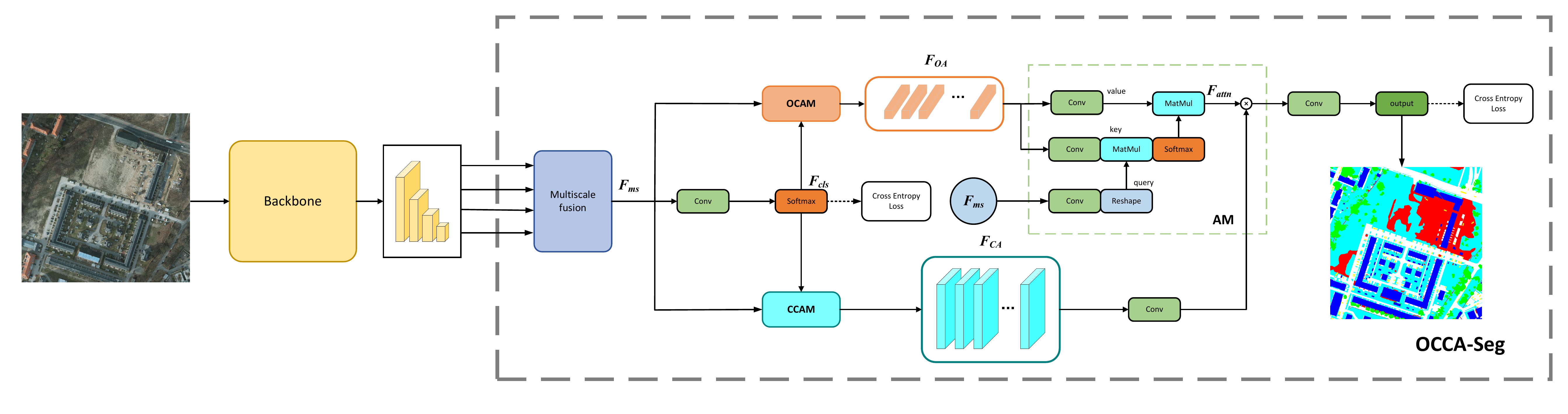}
    \caption{Overall architecture of OCCA-Seg}
    \label{fig:overall-architecture}
\end{figure*}
\subsection{Semantic Segmentation in Remote Sensing Images}
Semantic segmentation in remote sensing images is a fundamental task in computer vision, 
which aims to assign a semantic label to each pixel.
Traditional methods are based on handcrafted feature engineering and machine learning classifiers \cite{OBLCSC}.
These methods are limited by the reliance on handcrafted features and high labor costs. 
The handcrafted features have inadequate representation capabilities in various scenarios.
The classifiers based on machine learning methods have insufficient generalization ability, 
due to their limited number of parameters \cite{bagwari_comprehensive_2023}.
In recent years, various methods based on deep learning have been proposed, 
which significantly improved the performance of SSRSIs.
\par In the past decades, various CNN-based networks were applied for SSRSIs, 
such as FCN \cite{Long_2015_CVPR}, UNet\cite{10.1007.978_UNet}, SegNet\cite{7803544}, DeepLab \cite{7913730_DEEPLAB}, RAPNet \cite{9031697_RAPNet}, etc.
As mentioned in the introduction, the limited receptive field of CNN-based methods constrains their capability for long-range modeling.
There are three main paths to tackle this drawback: improving the feature map resolution, 
enlarging the convolutional kernel size, and exploiting the long-range context.
For the aspect of improving the feature map resolution, HRNet \cite{sun2019high} is proposed to maintain the feature-map resolution throughout the forward pass.
Semantic FPN \cite{10.1371/journal.pone.0275194} and other pyramid-based methods are proposed to enhance the feature representation by the fusion of multi-scale features, 
which helps the model preserve high-resolution features and more geometric details.
UNet \cite{10.1007.978_UNet} is designed to capture the shallow and high-resolution features by skip connections to use high-resolution feature maps.
For the aspect of enlarging the convolutional kernel size, Chen et al. \cite{7913730_DEEPLAB} proposed the DeepLab family 
to enlarge the receptive field by dilated convolutions.
DenseASPP\cite{yang2018denseaspp} densifies the dilated rates to cover larger scale receptive fields.
For the aspect of exploiting the long-range context, ParseNet \cite{liu2015parsenet} is proposed to combine global context 
information with local features by its contextual module. Yuan et al \cite{yuan2021ocnet} proposed OCNet to capture the long-range context by the object context module.
They modeled the dense relations between any two of all pixels via the combination of two sparse relation matrices by an efficient interlaced sparse self-attention mechanism.
Furthermore, OCRNet \cite{10.1007.978.11_OCR} computes the relations between each pixel
and each object category, augmenting the representation of each pixel with the object context.
\par In order to overcome the receptive field limitation, we focus on using attention mechanism to compute long-range context.
The attention mechanism in this research does not only compute the relationships between each pixel in different regions,
but also takes into account the relationships between the channels in each pixel and the object regions.

\subsection{Attention Mechanism}
To capture the relationships between features, most methods can be classified into two categories: 
transformer-based holistic architectures (TBHAs) and attention-based modular methods (ABMMs).
The TBHAs leverage self-attention mechanisms to model the relationships in internal parts of input features,
such as MaskFormer \cite{cheng2022masked}, SegVit \cite{zhang2024segvit}, HRVit \cite{gu2022hrvit}, CTSeg \cite{chen2025ctseg}, etc.
MaskFormer unifies various segmentation tasks through mask classification. SegVit leverages class tokens for semantic mask generation. 
HRVit integrates the high-resolution multi-branch architecture with ViT for fusing multi-scale features effectively.
CTSeg combines CNN and ViT to efficiently capture both local and global context for the segmentation task.
These architectures demonstrate the versatility of transformer-based methods for dense prediction tasks.
However, they are computationally expensive due to many matrix multiplications, limiting their applicability in the scenarios of SSRSIs.

\par In contrast, ABMMs, such as OCNet \cite{yuan2021ocnet}, OCRNet\cite{10.1007.978.11_OCR}, and D2LS \cite{Zou_2025_ICCV}, also play an important role in enhancing feature representations, 
which can guide the model to focus on certain areas and assign weights to different regions in the feature maps.
These methods can be easily integrated into existing networks and are generally more computationally efficient than transformer-based methods.
For example, both OCNet and OCRNet are integrated with HRNet \cite{sun2019high} for better accuracy of SSRSIs.
OCNet represents the dense relation matrix using two sparse matrices, so that it can efficiently represent the contextual features between pixels. 
OCRNet computes the object region representation by aggregating the representations of the pixels lying in the object region.
The D2LS method with ConvNeXt \cite{liu2022convnet} leverages a learnable dynamic dictionary and object category semantic information to compute
the cross-attention between features. 
Attention-based modular methods are increasingly popular because of their high computational efficiency and flexibility.

\par The OCCA-Seg proposed in this research is an attention-based modular method, which can be easily plugged into existing semantic segmentation frameworks. 
It captures the object-level and channel-wise contextual attention by OCAM and CCAM modules, respectively.
These features are then integrated for the final segmentation prediction. 
By jointly refining the pixel-level features through object-level and channel-wise context attention, OCCA-Seg can learn more class-discriminative representations,
thereby promoting intra-class feature consistency and inter-class feature separability among land-cover categories.

\section{Methodology}

\subsection{The Overall Architecture Design}
\begin{figure} 
    \centering
    \includegraphics[width=8cm]{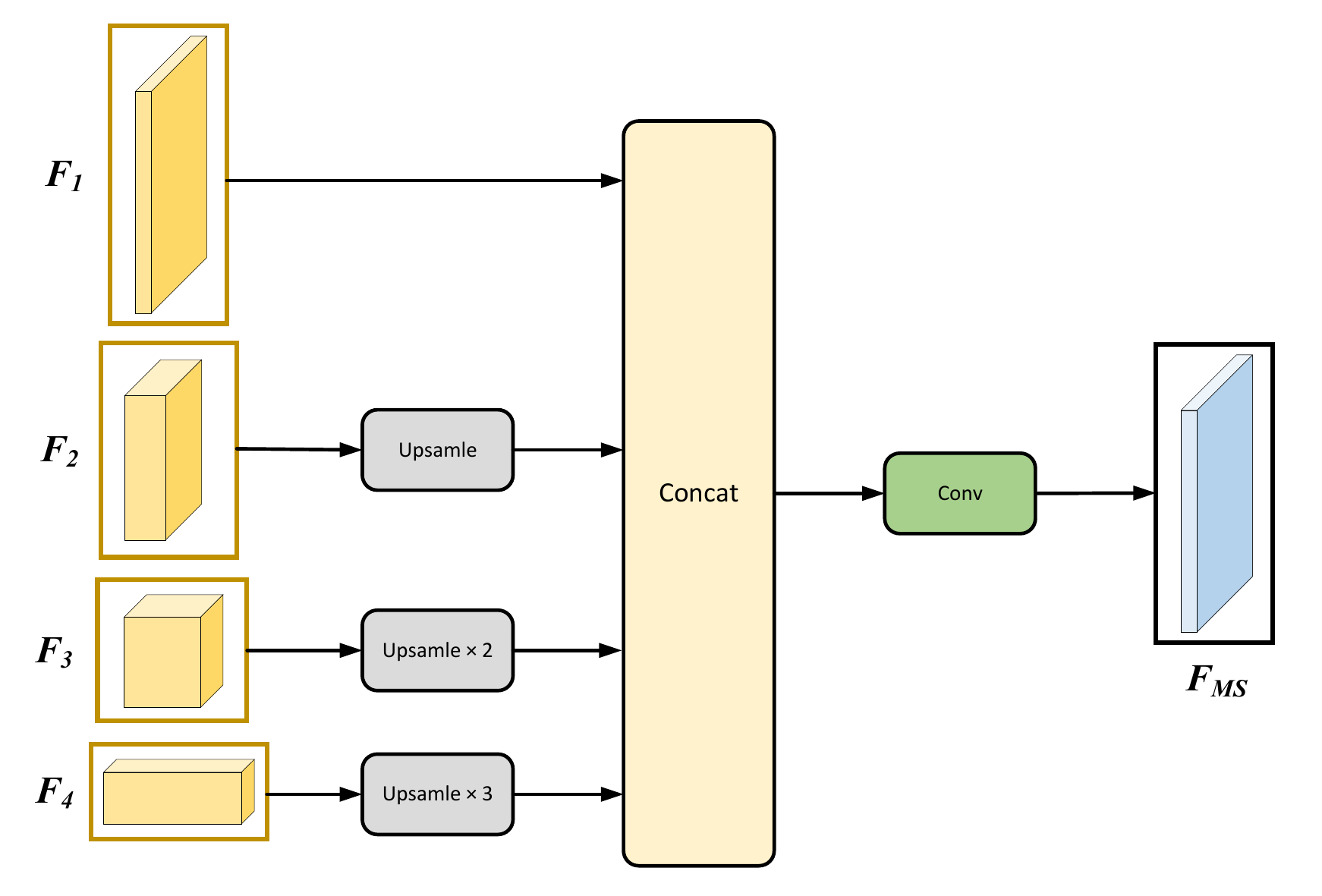}
    \caption{The architecture of MSF}
    \label{fig:msf}
\end{figure}

In order to effectively capture the contextual information and enlarge the receptive field, we propose the OCCA-Seg network, 
which integrates object and channel contextual attention mechanisms to enhance pixel-level feature representations.
The overall architecture of OCCA-Seg in Figure \ref{fig:overall-architecture} consists of three main components: OCAM, CCAM and AM.
Firstly, the multi-scale features from the backbone are fed into the multi-scale Fusion module (MSF) to aggregate the features at different scales and 
generate the unified scale feature map $\boldsymbol{F}_{ms} \in\mathbb{R}^{C \times H \times W}$, 
where $C$, $H$, and $W$ denote the number of channels, height and width of the feature map, respectively, as shown in Figure \ref{fig:msf} and Eq. \eqref{eq: msf}.
$\boldsymbol{F}_{ms}$ is used as the pixel representation in the OCCA-Seg network.

\begin{equation}
\boldsymbol{F}_{ms} = Stack(\boldsymbol{F}_1, f^{1}_{up}(\boldsymbol{F}_2), \dots, f^{\mathrm{M}-1}_{up}(\boldsymbol{F}_M)),
\label{eq: msf}
\end{equation}
\noindent The function $Stack(*)$ denotes the concatenation operation along the channel dimension, 
and $f^{i-1}_{up}(\boldsymbol{F}_{i})$ represents the upsampling operation with $i-1$ times for the $i$-th feature map.
$M$ denotes the total number of feature maps extracted from the backbone network. Here, $M$ is 4 in this work.
\par After the multi-scale fusion, the pipeline employs a dual-branch architecture: the OCAM and CCAM branches.
The OCAM is responsible for capturing object-level contextual information, 
while the CCAM focuses on channel-wise contextual dependencies. 
The outputs from both branches are then fused by the AM to generate the final enhanced feature map.
The global computation of OCCA-Seg can be summarized as follows:  
\begin{equation}
\begin{cases}
    \boldsymbol{F}_{OA} = \text{OCAM}(\boldsymbol{F}_{ms}, \boldsymbol{F}_{cls}) \\
    \boldsymbol{F}_{CA} = \text{CCAM}(\boldsymbol{F}_{ms}, \boldsymbol{F}_{cls}) \\
    \boldsymbol{F}_{AG} = \text{AM}(\boldsymbol{F}_{OA}, \boldsymbol{F}_{CA}, \boldsymbol{F}_{ms})
\end{cases}
\label{eq:occa-seg}
\end{equation}
Where $\boldsymbol{F}_{cls}\in\mathbb{R}^{C' \times H \times W}$ is defined as the class region representation. $C'$ denotes the number of classes.
$\boldsymbol{F}_{OA}\in\mathbb{R}^{C' \times C}$, $\boldsymbol{F}_{CA}\in\mathbb{R}^{C'' \times H \times W}$, and $\boldsymbol{F}_{AG}\in\mathbb{R}^{C \times H \times W}$ 
are the outputs of the OCAM, CCAM, and AM, respectively. 
$\boldsymbol{F}_{OA}$ is the object-level contextual representation.
$\boldsymbol{F}_{CA}$ is the channel-wise contextual representation. 
$\boldsymbol{F}_{AG}$ is the final enhanced feature map.
$C''$ is the product of $C$ and $C'$, described as follows.
\begin{equation}
C'' = C \cdot C'
\end{equation}
\par As shown in Figure \ref{fig:overall-architecture}, the AM is a key submodule that fuses the outputs of the OCAM and CCAM to generate the final enhanced feature map.
There are two stages in the AM. The first stage is the attention computation.
\begin{equation}
\operatorname{Attn}(\boldsymbol{Q},\boldsymbol{K},\boldsymbol{V})= \operatorname{softmax}\left(\frac{\boldsymbol{Q}\boldsymbol{K}}{\sqrt{d_k}}\right)\boldsymbol{V}.
\label{eq:attn}
\end{equation}
\par Inspired by the Transformer architecture, the first stage of the AM aims to capture the relationships between the unified pixel feature and the object-level context. 
This is achieved by applying a cross-attention mechanism (shown in Eq. \ref{eq:attn}) that allows each position in the $\boldsymbol{F}_{ms}$ to attend to each class representation of the $\boldsymbol{F}_{OA}$, 
effectively integrating contextual information from the combining information of each pixel feature and the object-level contextual representation, shown as Eq. \ref{eq:am-a}.
\begin{equation}
\begin{cases}
\boldsymbol{Q} = Conv^{\top}(\boldsymbol{F}_{ms}),\\
\boldsymbol{K} = Conv^{\top}(\boldsymbol{F}_{OA}),\\
\boldsymbol{V} = Conv(\boldsymbol{F}_{OA}),\\
\boldsymbol{F}_{\mathrm{attn}}= \operatorname{Attn}(\boldsymbol{Q},\boldsymbol{K},\boldsymbol{V})
\end{cases}
\label{eq:am-a}
\end{equation}
Where \begin{math}Conv(*)\end{math} and \begin{math} Conv^{\top}(*)\end{math} denote the convolution and transposed convolution, respectively.
The query $\boldsymbol{Q} \in \mathbb{R}^{N \times C}$ is convolved and reshaped from $\boldsymbol{F}_{ms}$ that denotes the pixel representation.
It can be interpreted that the query consists of $N$ tokens with $C$ dimension.
Both the key $\boldsymbol{K}\in\mathbb{R}^{C \times C'}$ and the value $\boldsymbol{V}\in\mathbb{R}^{C' \times C}$ are convolved and reshaped from $\boldsymbol{F}_{OA}$ that denotes the object-level contextual representation.
It can be interpreted that the key consists of $C$ tokens with $C'$ dimension, and the value consists of $C'$ tokens with $C$ dimension.
Just like the self-attention design in Transformer\cite{vaswani2023attentionneed}, our computation of the attention $\boldsymbol{F}_{attn}\in\mathbb{R}^{N \times C}$ can be described as mapping the query from $\boldsymbol{F}_{ms}$ and a set of key-value pairs from $\boldsymbol{F}_{OA}$ to an output $\boldsymbol{F}_{attn}$, where the query, key, and value are all vectors.
The output is computed as a weighted sum of the values, where the weight assigned to each value is computed by a compatibility function of the query with the corresponding key.

\par The second stage of the AM focuses on integrating the channel-wise context from the CCAM and $\boldsymbol{F}_{ms}$. 
This is achieved by applying a channel attention mechanism that allows each element in the $\boldsymbol{F}_{attn}$ to attend to the corresponding element of the $\boldsymbol{F}_{CA}$, 
effectively enhancing each pixel representation with channel-wise context. 
In this stage, the channel-wise context representation $\boldsymbol{F}_{CA}$ is used to reweight the attention-enhanced feature $\boldsymbol{F}_{attn}$ by element-wise multiplication, emphasizing the contribution of each channel in $\boldsymbol{F}_{CA}$ to the prediction.
\begin{figure} 
    \centering
    \includegraphics[width=8cm]{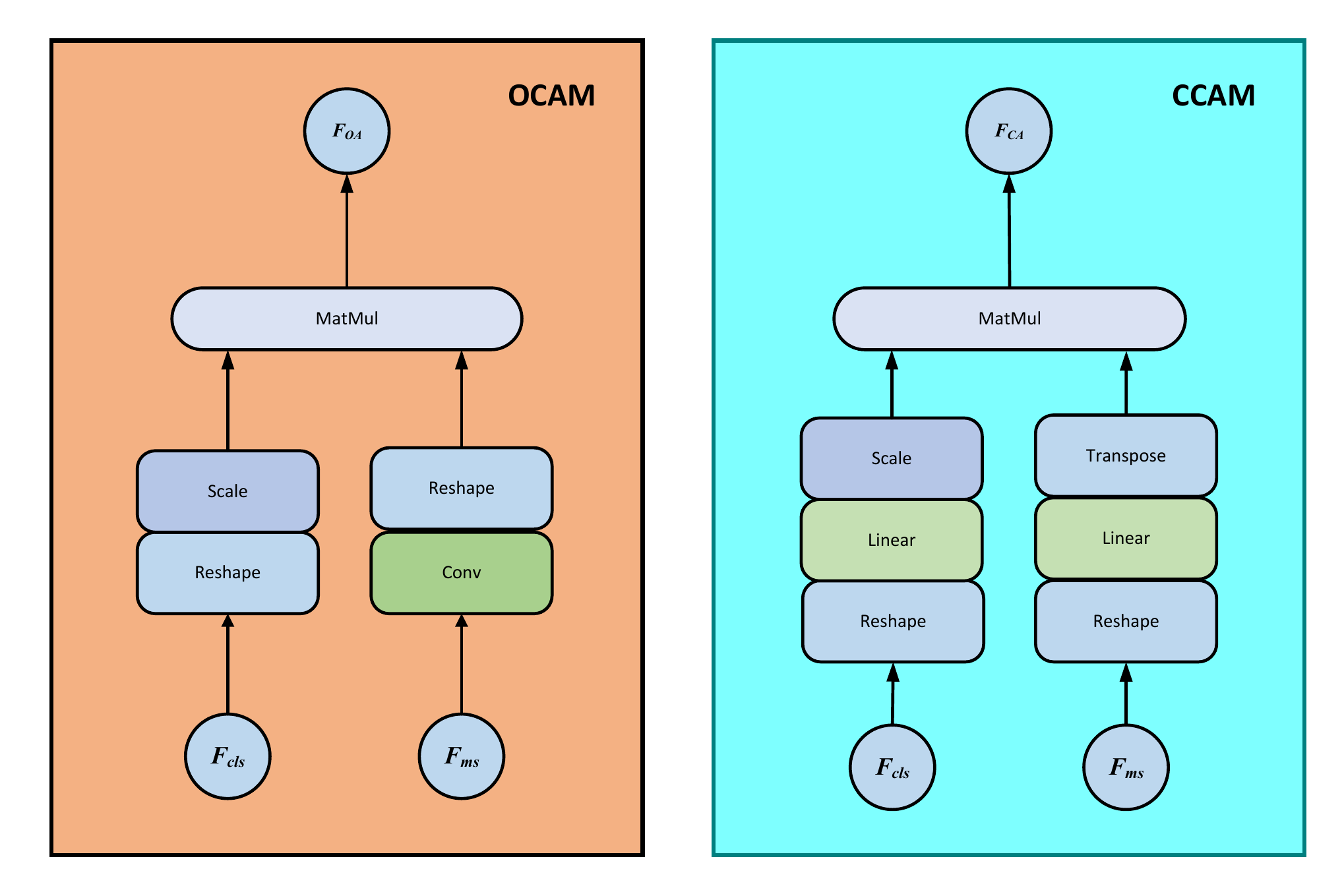}
    \caption{The architecture of OCAM and CCAM}
    \label{fig:ocam-ccam}
\end{figure}

\begin{equation}
    \boldsymbol{F}_{AG} = Conv(\boldsymbol{F}_{CA}) \odot \operatorname{Reshape}(\boldsymbol{F}_{\mathrm{attn}})
    \label{eq:aggreator-b}
\end{equation}
Here, \begin{math}\odot\end{math} denotes the element-wise multiplication.
We reshape $\boldsymbol{F}_{\mathrm{attn}}$ from $(N, C)$ to $(C, H, W)$ to match the shape of projected $\operatorname{Conv}(\boldsymbol{F}_{CA})$.
Simultaneously, we convert the number of channels of $\boldsymbol{F}_{CA}$ from $C''$ to $C$ using a convolution layer for channel alignment and reweighting $\boldsymbol{F}_{\mathrm{attn}}$.
\par Finally, $\boldsymbol{F}_{AG}$ is processed by a convolution layer to produce the prediction map $\boldsymbol{F}_{pred}\in\mathbb{R}^{C' \times H \times W}$.
\par In summary, OCCA-Seg first aggregates multi-scale features to obtain the scale-unified pixel representation.
The OCAM and CCAM then capture the object-level and channel-wise contextual information, respectively. 
Finally, the AM makes fusion of these contextual information with the pixel representation to produce context-enhanced features for SSRSIs. 
The overall design of OCCA-Seg enables the network to exploit contextual information and enlarge the receptive field.

\subsection{Object Contextual Attention Module}
\par As shown in the left part of Figure \ref{fig:ocam-ccam}, the OCAM is designed to capture object-level contextual information by modeling the relationships between different class regions and the pixel representation.
The computation of the OCAM involves generating an object attention map that highlights the relationships between each class region and each pixel, as shown in Eq. \ref{eq:ocam}. 
It is worth noting that the relationships between each class region and each pixel is calculated by matrix multiplication.
This operation makes these relationships aggregated over all pixels for each class region representation that does not retain the spatial representation of individual pixels but rather summarize it into $\boldsymbol{F}_{OA}$ in a global format.

\begin{equation}
    \begin{cases}
        \boldsymbol{F'}_{\mathrm{ms}} =\operatorname{Reshape}(Conv(\boldsymbol{F}_{ms}))\\
        \boldsymbol{F'}_{\mathrm{cls}} = \operatorname{Reshape}(\boldsymbol{F}_{\mathrm{cls}})\\
        \boldsymbol{F}_{\mathrm{OA}} =  Scale \cdot \boldsymbol{F'}_{cls} \times \boldsymbol{F'}_{ms}
    \end{cases}
    \label{eq:ocam}
\end{equation}
\par Here, \begin{math}\times\end{math} denotes matrix multiplication. 
\begin{math}\cdot\end{math} denotes scalar multiplication.
$Scale$ is a scaling factor to control magnitude of the $\boldsymbol{F}_{cls}$, and is set to 1 in this research.
Given $\boldsymbol{F}_{ms}$ and $\boldsymbol{F}_{cls}$, 
the OCAM establishes associations between semantic regions and pixel-level features.
As expressed in Eq. \ref{eq:ocam}, $\boldsymbol{F'}_{ms}$ is obtained by a convolution and reshape operation that transforms the shape of $\boldsymbol{F}_{ms}$ from $(C, H, W)$ to $(N, C)$, where $N$ is the number of pixels of the feature map.
Similarly, $\boldsymbol{F'}_{cls}$ is obtained by reshaping $\boldsymbol{F}_{cls}$ into a matrix of shape $(C', N)$.
Finally, we can compute the object-level contextual representation $\boldsymbol{F}_{OA}$ by multiplying $\boldsymbol{F'}_{cls}$ with $\boldsymbol{F'}_{ms}$ and scaling the result by $Scale$.
The shape of $\boldsymbol{F}_{OA}$ is $(C', C)$ after the matrix multiplication. Then, $\boldsymbol{F}_{OA}$ is fed into the AM for subsequent aggregation.

\par The pixel representation $\boldsymbol{F'}_{ms}$ is combined with the class region representation $\boldsymbol{F'}_{cls}$ through matrix multiplication to produce the object-level contextual representation $\boldsymbol{F}_{OA}$.
In this way, features from spatially distant locations can contribute to the representation of the same semantic class, allowing the module to capture long-range object context beyond the local receptive fields of convolutional layers.
The resulting $\boldsymbol{F}_{OA}$ encodes object-level contextual information and is passed to the AM together with the pixel representation $\boldsymbol{F}_{ms}$. 
This enables the AM to incorporate object-level cues into pixel representations, which can help distinguish categories with similar local appearances and improve feature consistency among pixels belonging to the same object category. 
By leveraging the combination of the pixel representation and the class region information, 
the OCAM provides complementary contextual cues for semantic segmentation.

\subsection{Channel Contextual Attention Module}
\par As shown in the right part of Figure \ref{fig:ocam-ccam}, the CCAM is designed to capture channel-wise contextual information 
by modeling the relationships between the channels of each pixel and class region representations.
The computation of CCAM involves generating a channel attention map that highlights the importance of each pixel channel for different class region representation. 
This attention map is then used to reweight the feature channels by class region representation to enhance the channel-wise feature representation.

\begin{equation}
    \begin{cases}
        \boldsymbol{F''}_{\mathrm{cls}} = \operatorname{Linear}(\operatorname{Reshape}(\boldsymbol{F}_{\mathrm{cls}}))\\
        \boldsymbol{F''}_{\mathrm{ms}} = \operatorname{Linear^\top}(\operatorname{Reshape}(\boldsymbol{F}_{\mathrm{ms}}))\\
        \boldsymbol{F}_{\mathrm{CA}} = Scale \cdot \boldsymbol{F''}_{\mathrm{cls}} \times \boldsymbol{F''}_{\mathrm{ms}}
    \end{cases}
    \label{eq:ccam}
\end{equation}
Here, the operator $\operatorname{Reshape}$ reshapes the feature maps into a suitable form for linear transform. 
$\boldsymbol{F''}_{\mathrm{cls}} \in \mathbb{R}^{N \times C' \times 1}$ is obtained from $\boldsymbol{F}_{\mathrm{cls}}$ by a reshape and linear transform.
$\boldsymbol{F''}_{\mathrm{ms}} \in \mathbb{R}^{N \times 1 \times C}$ is obtained from $\boldsymbol{F}_{\mathrm{ms}}$ by a cascade of reshape, linear transform, and transpose.
$\operatorname{Linear}$ denotes a learnable linear transform. 

The resulting $\boldsymbol{F}_{\mathrm{CA}}$ is computed by the matrix multiplication of the class region features $\boldsymbol{F''}_{\mathrm{cls}}$ and the pixel representation $\boldsymbol{F''}_{\mathrm{ms}}$, representing the channel-wise attention-enhanced feature map.
The original shape of $\boldsymbol{F}_{\mathrm{CA}}$ is $(N, C', C)$. 
For easy aggregation in the AM, we reshape it to $(C'', H, W)$.
\par $\boldsymbol{F}_{\mathrm{CA}}$ captures relationships between pixel feature channels and class region representations, providing channel-wise context. 
The AM uses it to weight the $\boldsymbol{F}_{\mathrm{attn}}$ and emphasize channels that are informative for each class. 
Together, the object-level context from OCAM and the channel-wise cues from CCAM produce more discriminative feature representations.

\section{Experiments}
\subsection{Datasets}
In this research, we evaluate OCCA-Seg on four remote sensing datasets for semantic segmentation tasks, which cover different spatial resolutions, scene complexities, and land-cover types.
These datasets include LoveDA\cite{wang2021loveda}, ISPRS Potsdam\cite{isprs2024potsdam}, ISPRS Vaihingen\cite{isprs2024vaihingen}, and UAVid\cite{LYU2020108} dataset.
For the splitting and preprocessing of these datasets, we follow the same protocol as SFA-Net\cite{hwang2024sfa} and D2LS\cite{zou2025dynamic}. 
Through experiments on these datasets, we are able to comprehensively validate the adaptability of the proposed method in different remote sensing scenarios.

\subsection{Implementation Details}
All experiments in this research are conducted on an NVIDIA GeForce 4090 GPU with 24GB memory.
And all codes are implemented in Python using the PyTorch framework.
For ablation experiments and comparison with other state-of-the-art methods, 
we use the different implementation settings.

\par For ablation experiments, we use a YOLO26 model pretrained on ImageNet dataset as our baseline model.
In the training, we adopt the AdamW optimizer with weight decay for network optimization.
The initial learning rate is set to $0.001$ and the weight decay is set to $0.0005$.
We apply warmup mechanism to prevent gradient explosion.
The number of warmup epochs, momentum and bias learning rate are set as 3.0, 0.8 and 0.1 respectively.
\par For comparison with SOTA methods, we use the ConvNext-Base pretrained on ImageNet dataset as our backbone model.
 The initial learning rate is set to $8 \times 10^{-5}$ and the weight decay is set to $0.01$.
 The warmup mechanism is not applied for this setting.

\subsection{Ablation experiments}
For evaluation of OCCA-Seg, we conduct ablation experiments on the ISPRS Potsdam and Vaihingen datasets.
YOLO26n-SEM from ultralytics is used as the baseline model.
Both Potsdam and Vaihingen datasets have the same categories, including impervious surfaces (IS), buildings (BL), low vegetation (LV), trees (TE), and cars (CA).
For each class of ISPRS Potsdam and Vaihingen datasets, the IoU, F1-Score(mF1) and pixel accuracy (OA) metrics are used as the main evaluation metrics and shown in Table \ref{tab:table1}.
The families of YOLO are famous in the field of computer vision. The semantic segmentation capability is supported by the YOLO26-SEM model.
We replace the head of YOLO26n-SEM with our OCCA-Seg for improving the semantic segmentation performance in ablation experiments.

\begin{table*}[!t]
\caption{Performance comparison on the ISPRS Potsdam and Vaihingen dataset}
\normalsize
\centering
    \setlength{\tabcolsep}{1.5mm}{
    \begin{tabular}{ccccccccccccc} 
        \hline
        \cline{2-8}   
        &Method &Dataset &IS(\%) &BL(\%) &LV(\%) &TE(\%) &CA(\%) &mIoU(\%) &mF1(\%) &OA(\%)\\
        \hline
        &YOLO26n-SEM  &Potsdam &81.4 &87.6 &71.5 &70.6 &79.2 &78.1 &88.0 &87.5 \\
        &YOLO26n-OCCA  &Potsdam  &84.0 &91.2 &73.2 &74.4 &83.3 &81.2 &89.8 &89.5 \\
        &YOLO26n-SEM  &Vaihingen &79.8 &85.7 &64.4 &74.0 &59.1 &72.6 &83.7 &86.4 \\
        &YOLO26n-OCCA  &Vaihingen &80.6 &87.3 &66.1 &74.8 &64.8 &74.7 &85.3 &87.2  \\
        \hline
    \end{tabular}}
 \label{tab:table1}
\end{table*}
\par As shown in Table \ref{tab:table1}, the proposed OCCA head consistently improves segmentation performance on both the ISPRS Potsdam and Vaihingen datasets. 
On Potsdam, YOLO26n-OCCA increases mIoU, mF1, and OA from 78.1, 88.0, and 87.5 to 81.2, 89.8, and 89.5, respectively. 
On Vaihingen, the corresponding metrics improve from 72.6, 83.7, and 86.4 to 74.7, 85.3, and 87.2. 
The IoU also increases for every category on both datasets, with the largest gain on Vaihingen observed for cars (from 59.1 to 64.8).
These results of ablation experiments indicate that the OCCA-Seg provides consistent benefits across different datasets, supporting the effectiveness of aggregating object-level and channel-wise context into the baseline model.

\subsection{Comparison with SOTA Methods}
To evaluate the capability of our proposed OCCA-Seg method in SSRSIs, we conducted comprehensive experiments across multiple datasets to compare it with other state-of-the-art methods.
\par \textbf{Potsdam Result.} 
The ISPRS Potsdam dataset has 38 items in total. In this comparison, the ISPRS official dataset preprocessing scheme is adopted, where the training subset contains 
2-10, 2-11, 2-12, 3-10, 3-11, 3-12, 4-10, 4-11 4-12, 5-10, 5-11, 5-12, 6-10, 6-11, 6-12, 6-7, 6-8, 6-9, 7-10, 7-11, 7-12, 7-7, 7-8 and 7-9,
including 24 items in total.
The validation subset contains the remaining items, including 14 items in total.
Both image and mask of each item are split into 1024$\times$1024 patches.
During the test, we adopt the multiscale data augmentation strategy with the scale of (0.5, 1.0, 1.5, 2.0).
Table \ref{tab:table2} illustrates the results of the SOTA methods on the preprocessed ISPRS Potsdam dataset,
where the OCCA-Seg method obtains 87.51\% mIoU, 93.21\% mF1, and 91.85\% OA.
It can be observed that OCCA-Seg outperforms other methods listed in Table \ref{tab:table2} for overall performance.
At the category level, OCCA-Seg achieves the best IoU for impervious surfaces (88.80\%), trees (81.23\%), and cars (93.61\%).
Its IoU on buildings is also competitive, which is closed to the best results of 94.83\% achieved by D2LS\cite{zou2025dynamic}.
For the category of low vegetation, OCCA-Seg achieves an IoU of 79.13\%, ranking fourth among the compared methods.
This result is lower than the best IoU of 79.66\% achieved by EfficientVit\cite{cai2023efficientvit} by 0.53 percentage points.
In summary, the OCCA-Seg achieves the leading performance across different categories on the ISPRS Potsdam dataset.
\begin{table*}[!t]
\caption{Performance comparison on the ISPRS Potsdam dataset}
\normalsize
\centering
    \setlength{\tabcolsep}{1.5mm}{
    \begin{tabular}{cccccccccccccc} 
        \hline
        &Method &IS(\%) &BL(\%) &LV(\%) &TE(\%) &CA(\%) &mIoU(\%) &mF1(\%) &OA(\%)\\
        \hline
        &FarSeg\cite{zheng2020foreground}    &85.28 &91.33 &76.46 &77.36 &91.37 &84.36 &91.21 &89.87 \\
        &FarSeg++\cite{10188509}    &87.77 &93.88 &77.21 &79.64 &92.65 &86.23 &91.99 &91.22 \\
        &ISNet\cite{jin2021isnet}    &87.53 &93.68 &78.60 &80.03 &93.06 &86.58 &92.67 &91.27 \\
        &UNetFormer\cite{wang2022unetformer}   &87.58 &94.23 &79.45 &80.18 &91.27  &86.54 &92.77 &91.46 \\
        &EfficientVit\cite{cai2023efficientvit}   &84.37 &91.49 &79.66 &77.43 &88.20  &84.23 &90.11 &89.61 \\
        &PEM\cite{cavagnero2024pem}   &86.86 &93.68 &77.34 &79.65 &90.23  &85.55 &91.73 &90.48 \\
        &LOGCAN++\cite{10884928}   &83.98 &91.89 &72.57 &80.15 &75.32   &82.06 &89.94 &91.47 \\
        &MSEONet\cite{huang2025mseonet}   &84.12 &91.45 &72.91 &75.62 &83.37  &81.33 &84.17 &88.68 \\
        &D2LS\cite{zou2025dynamic}  &88.16 &94.83 &78.24 &80.77 &93.23 &87.04 &92.94 &91.73 \\
        &OCCA-Seg(ours)  &88.80 &94.77 &79.13 &81.23 &93.61  &87.51 &93.21 &91.85  \\
        \hline
    \end{tabular}}
\label{tab:table2}
\end{table*}

\par \textbf{Vaihingen Result.}
The ISPRS Vaihingen dataset has 33 items in total. In this comparison, 
we used IDs: 1, 3, 5, 7, 11, 13, 15, 17, 21, 23, 26, 28, 30, 32, 34, and 37 for training.
And the remaining 17 items are used for validation.
Both image and mask of each item are split into 1024$\times$1024 patches.
During the test, we adopt the single scale data augmentation strategy with the scale of 1.0.
As shown in Table \ref{tab:vaihingen}, the OCCA-Seg achieves the best performance among the compared methods on the ISPRS Vaihingen dataset.
The mean of IoU, F1-Score, and overall accuracy are 84.03\%, 91.17\%, and 92.44\%, respectively.
For the class level evaluation, it ranks first in IoU for impervious surfaces (91.67\%).
\begin{table*}[!t]
\caption{Performance comparison on the ISPRS Vaihingen dataset}
\normalsize
\centering
\setlength{\tabcolsep}{1.5mm}{%
\begin{tabular}{cccccccccccccc}
\hline
&Method &IS(\%) &BL(\%) &LV(\%) &TE(\%) &CA(\%) &mIoU(\%) &mF1(\%) &OA(\%)\\
\hline
&FarSeg\cite{zheng2020foreground}        & 84.10 & 89.27 & 71.10 & 80.10 & 71.1  & 79.14 & 87.88 & 89.57 \\
&FarSeg++\cite{10188509}      & 86.47 & 90.86 & 74.39 & 82.87 & 76.88 & 82.29 & 89.77 & 90.91 \\
&ISNet\cite{jin2021isnet}         & 85.91 & 90.89 & 72.96 & 81.93 & 80.01 & 82.36 & 90.19 & 90.52 \\
&UNetFormer\cite{wang2022unetformer}    & 88.14 & 91.47 & 71.23 & 81.46 & 79.22 & 82.30 & 90.31 & 90.82 \\
&ConvNeXt\cite{liu2022convnet}      & 87.13 & 91.56 & 73.81 & 83.48 & 78.37 & 82.87 & 90.50 & 91.36 \\
&EfficientViT\cite{cai2023efficientvit}  & 87.26 & 90.42 & 69.14 & 78.41 & 77.37 & 80.52 & 87.56 & 89.41 \\
&DOCNet\cite{ma2024docnet}        & 87.76 & 91.36 & 74.45 & 82.68 & 82.13 & 83.73 & 91.11 & 91.36 \\
&D2LS\cite{zou2025dynamic}           & 87.07 & 91.33 & 72.9 & 82.15 & 80.90 & 82.87 & 90.50 & 91.02 \\
&OCCA-Seg(ours) & 91.67 & 91.57 & 73.89 & 82.37 & 80.64 & 84.03 & 91.17 & 92.44 \\
\hline
\end{tabular}
\label{tab:vaihingen}
}
\end{table*}

\par \textbf{LoveDA Result.}
The LoveDA dataset is designed for SSRSI competition, only accepting online tests.
In order to obtain a better performance, we put the training and validation sets together for training our model.
We report the comparison results from the official LoveDA online test server at Table \ref{tab:loveda}.
As shown in this table, the proposed OCCA-Seg achieves the best performance among all compared methods on the LoveDA dataset.
It achieves the highest mIoU of 55.5\% on the LoveDA test set, exceeding D2LS\cite{zou2025dynamic} by 0.2 percentage point.
At category level, OCCA-Seg obtains the best IoU for water (82.2\%) and agriculture (66.0\%).
Compared with D2LS\cite{zou2025dynamic}, it also achieves gains in background, building, and road, 
while its IoU for barren and forest is lower by 0.6 and 0.4 percentage points, respectively.
These results indicate that OCCA-Seg achieves the best overall performance among the compared methods.

\begin{table*}[!t]
\caption{Performance comparison on the LoveDA dataset}
\normalsize
\centering
    \setlength{\tabcolsep}{1.5mm}{
    \begin{tabular}{ccccccccccccc} 
        \hline
        \cline{2-8}   
        &Method &\makecell{Background\\(\%)} &\makecell{Building\\(\%)} &\makecell{Road\\(\%)} &\makecell{Water\\(\%)} &\makecell{Barren\\(\%)} &\makecell{Forest\\(\%)} &\makecell{Agriculture\\(\%)} &\makecell{mIoU\\(\%)}\\
        \hline
        &TransUNet\cite{chen2024transunet}   &43.0 &56.1 &53.7 &78.0 &9.3 &44.9 &56.9 &48.9 \\
        &DC-Swin\cite{wang2022novel}   &41.3 &54.5 &56.2 &78.1 &14.5 &47.2 &62.4 &50.6  \\
        &UNetFormers\cite{wang2022unetformer} &44.7 &58.8 &54.9 &79.6 &20.1 &46.0 &62.5 &52.4 \\ 
        &AerialFormer\cite{hanyu2024aerialformer} &47.8 &60.7 &59.3 &81.5 &17.9 &47.9 &64.0 &54.1 \\
        &SFA-Net\cite{hwang2024sfa} &48.4 &60.3 &59.1 &81.9 &24.1 &46.2 &64.0 & 54.9 \\
        &MIFNet\cite{Xie10749979}   &49.6 &56.3 &50.3 &58.9 &25.7 &34.2 &52.7 &46.8 \\
        &MSEONet\cite{huang2025mseonet}   &54.6 &63.7 &53.5 &65.6 &39.3 &39.3 &54.7 &52.9 \\
        &BRSMamba\cite{huang2025multi}   &55.5 &64.8 &56.9 &71.4 &28.1 &42.1 &62.1 &54.4 \\
        &D2LS\cite{zou2025dynamic}   &47.6 &61.2 &59.1 &81.6 &23.8 &48.8 &64.8 &55.3\\
        &OCCA-Seg(ours)  &48.2 &61.3 &59.2 &82.2 &23.2 &48.4 &66.0 &55.5 \\
        \hline
    \end{tabular}}
 \label{tab:loveda}
\end{table*}

\par \textbf{UAVid Result.}
Unlike LoveDA and the ISPRS Potsdam and Vaihingen datasets, UAVid consists of video sequences captured by unmanned aerial vehicles (UAVs), posing challenges such as complex urban scenes and small moving objects.
Table~\ref{tab:uavid_segmentation} compares OCCA-Seg with representative methods on UAVid.
According to the reported mIoU, OCCA-Seg achieves 68.9\%, ranking second among the compared methods, 1.5 percentage points below SFA-Net (70.4\%).
At the class level, OCCA-Seg obtains the highest IoU for trees (81.4\%), vegetation (65.6\%), and humans (31.6\%), exceeding SFA-Net by 0.6, 1.0, and 0.9 percentage points, respectively.
However, it trails SFA-Net on clutter, buildings, roads, moving cars, and static cars, with the largest gaps on the two car classes (5.7 and 6.3 percentage points, respectively).
These results show that OCCA-Seg performs competitively on UAVid and is particularly effective for vegetation and human classes, while identifying vehicle segmentation as an area for further improvement.

\begin{table*}[!t]
  \centering
  \setlength{\tabcolsep}{1.5mm}
  \caption{Comparison of semantic segmentation performance on the UAVid dataset.}
  \begin{tabular}{lccccccccc}
    \hline
    Method &\makecell{Clutter\\(\%)} &\makecell{Building\\(\%)} &\makecell{Road\\(\%)} &\makecell{Tree\\(\%)} &\makecell{Vegetation\\(\%)} &\makecell{Moving Car\\(\%)} &\makecell{Static Car\\(\%)} &\makecell{Human\\(\%)} &\makecell{mIoU\\(\%)} \\
     \hline
    DANet\cite{Fu_2019_CVPR}  & 64.9 & 58.9 & 77.9 & 68.3 & 61.5 & 59.6 & 47.4 & 9.1 & 55.9 \\
    ABCNet\cite{li2021abcnet}  & 67.4 & 86.4 & 81.2 & 79.9 & 63.1 & 69.8 & 48.4 & 13.9 & 63.8 \\
    BANet\cite{wang2021transformer}  & 66.7 & 85.4 & 80.7 & 78.9 & 62.1 & 69.3 & 52.8 & 21.0 & 64.6 \\
    SegFormer\cite{NEURIPS2021_SEGFORMER}  & 66.6 & 86.3 & 80.1 & 79.6 & 62.3 & 72.5 & 52.5 & 28.5 & 66.0 \\
    UNetFormer\cite{wang2022unetformer}  & 68.4 & 87.4 & 81.5 & 80.2 & 63.5 & 73.6 & 56.4 & 31.0 & 67.8 \\
    SFA-Net\cite{hwang2024sfa}  & 70.2 & 89.0 & 82.7 & 80.8 & 64.6 & 77.5 & 67.5 & 30.7 & 70.4 \\
    OCCA-Seg(ours) & 69.0 & 88.9 & 81.7 & 81.4 & 65.6 & 71.8 & 61.2 & 31.6 & 68.9 \\
    \hline
  \end{tabular}
  \label{tab:uavid_segmentation}
\end{table*}

\par \textbf{Visualization Result.}
\begin{figure*}[ht]
  \centering
  \includegraphics[width=0.9\textwidth]{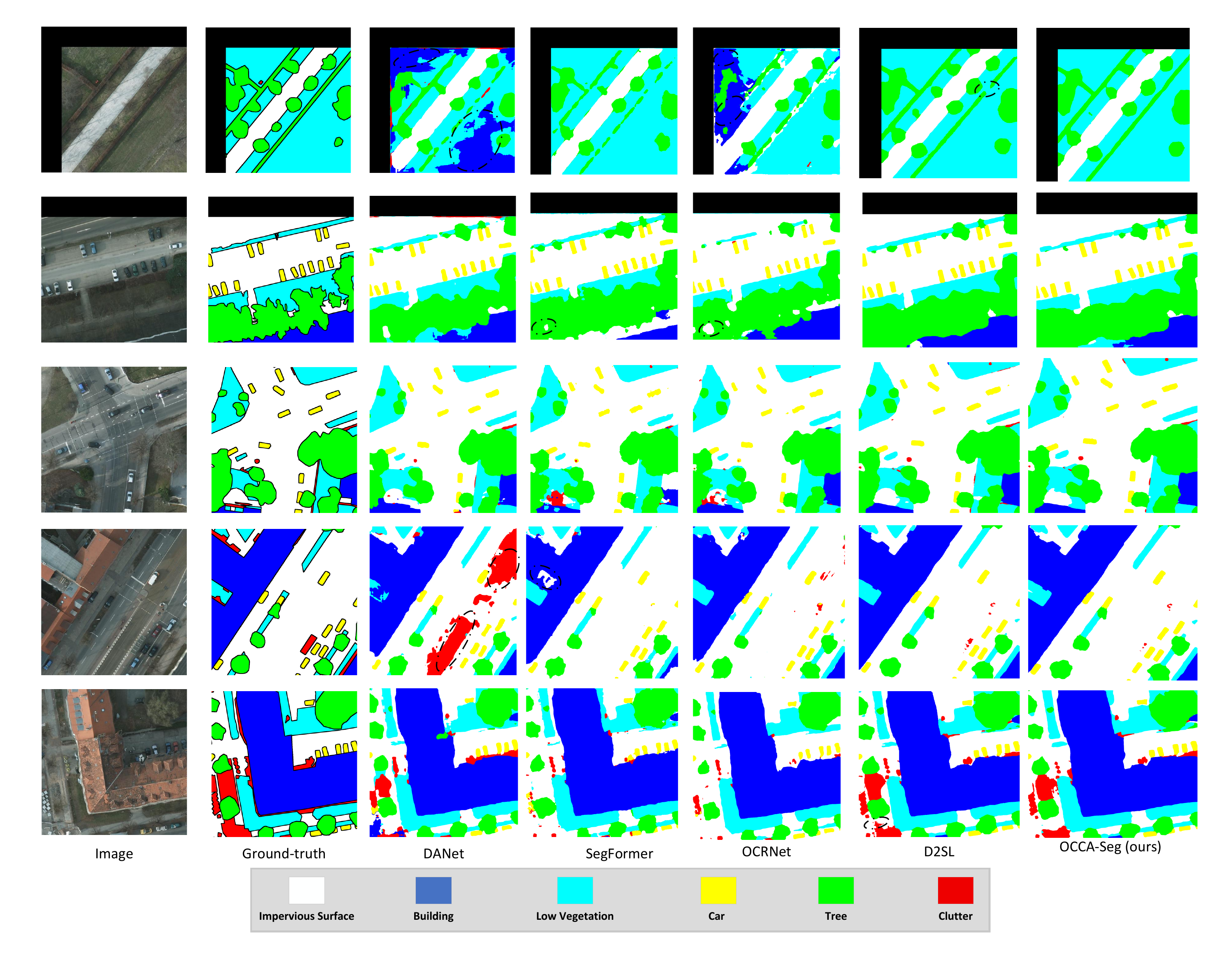}
  \caption{Visualization results of SOTA methods on ISPRS Potsdam dataset.}
  \label{fig:visualization_result}
\end{figure*}

In this research, the visualization experiments are performed on the ISPRS Potsdam dataset and shown in Figure~\ref{fig:visualization_result}.
As observed from these visualization results, the methods of DANet and OCRNet suffer from the widespread misclassification.
Some large voids will appear in the segmentation results of the Segformer method.
For the D2LS, some fracture points can be observed in the thin and long segmented objects, such as roads, rivers, and green belts.
Besides, it may produce the glitch noise on the edge of segmented objects.
In contrast, the proposed OCCA-Seg method demonstrates more accurate segmentation results, effectively mitigating the issues observed in other methods.

\section{Conclusion}
In this paper, we propose the OCCA-Seg framework to address the limitations of the receptive field and improve the distinguishing representation of the object context for SSRSIs.
Our method effectively captures both object-level and channel-wise contextual dependencies through a dual-branch architecture with attention mechanism:
the Object Contextual Attention Module (OCAM) captures category-level context, while the Channel Contextual Attention Module (CCAM) models channel-wise dependencies conditioned on class regions.
Their outputs are integrated with the multi-scale features by an AM to produce the final enhanced representation.
This design provides a means of incorporating both object-level and channel-wise contextual information to enhance the receptive field and pixel discrimination.

\section*{Acknowledgments}
This work is supported by the 2024 Ministry Provincial Cooperation Project of the Ministry of Natural Resources of the People's Republic of China (No.2024ZRBSHZ055), 
and research on the Integrated Protection and Adaptive Utilization Management Platform and Application for Cultural and Natural Heritage.


\bibliographystyle{IEEEtran}
\bibliography{IEEEabrv, ref}

\newpage

\vspace{11pt}

\begin{IEEEbiography}[{\includegraphics[width=1in,height=1.25in,clip,keepaspectratio]{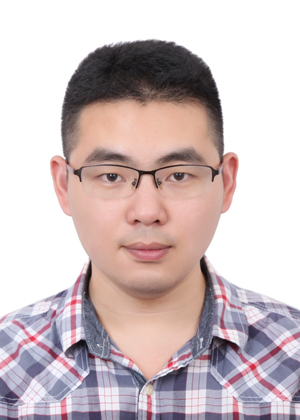}}]{Gang Yang}
received his B.S. and M.S. degrees from Beijing Jiaotong 
University in 2014 and 2019, respectively. He is working for Beijing Institute 
of Survey and Mapping and Beijing Key Laboratory of Urban Spatial 
Information Engineering, Beijing, China. His research interests include 
machine learning and remote sensing image processing.
\end{IEEEbiography}

\vspace{11pt}

\begin{IEEEbiographynophoto}{Si Chen}
received her master's degree in Urban Planning and Design from Beijing Institute of Civil Engineering and Architecture. 
In 2021, she was awarded the title of Senior Engineer at the Beijing Institute of Surveying and Mapping. 
Her research interests include conservation of historical and cultural cities, geographic information systems (GIS), and realistic 3D geospatial representation.
\end{IEEEbiographynophoto}

\begin{IEEEbiographynophoto}{Quan Zhou}
received his B.S. degree from Capital Normal University, China,
in 2015 and his M.S. degree from the University of Twente, Netherlands in
2018. Since 2018 he has been an engineer at Beijing Institute of Surveying
and Mapping. His research interests are in smart city, remote sensing, and
system architecture design.
\end{IEEEbiographynophoto}

\begin{IEEEbiographynophoto}{Jiangchuan Li}
received his B.S. and M.S. degrees from Beijing University of 
Civil Engineering and Architecture, China, in 2019 and 2022, respectively. 
Since 2022, he has been an assistant engineer at Beijing Institute of Surveying 
and Mapping. His research interests are in GIS and remote sensing.
\end{IEEEbiographynophoto}

\vfill

\end{document}